\documentclass[final]{cvpr}

\usepackage{times}
\usepackage{epsfig}
\usepackage{graphicx}
\usepackage{amsmath}
\usepackage{amssymb}
\usepackage[flushleft]{threeparttable} 
\usepackage{booktabs} 
\usepackage{multirow}
\usepackage{algorithm}
\usepackage{algpseudocode}

\DeclareMathOperator*{\argmax}{arg\,max}
\DeclareMathOperator*{\argmin}{arg\,min}

\usepackage[pagebackref=true,breaklinks=true,colorlinks,bookmarks=false]{hyperref}

\def\cvprPaperID{7847} 
\def\confYear{CVPR 2021}

\begin{document}

\title{Auto-Agent-Distiller: Towards Efficient Deep Reinforcement Learning Agents via Neural Architecture Search}

\author{Yonggan Fu\\
Rice University\\
{\tt\small yf22@rice.edu}
\and
Zhongzhi Yu\\
Rice University\\
{\tt\small zy42@rice.edu }
\and
Yongan Zhang\\
Rice University\\
{\tt\small yz87@rice.edu }
\and
Yingyan Lin\\
Rice University\\
{\tt\small yingyan.lin@rice.edu }
}

\maketitle

\begin{abstract}
AlphaGo's astonishing performance has ignited an explosive interest in developing deep reinforcement learning (DRL) for numerous real-world applications, such as intelligent robotics. However, the often prohibitive complexity of DRL stands at the odds with the required real-time control and constrained resources in many DRL applications, limiting the great potential of DRL powered intelligent devices. While substantial efforts have been devoted to compressing other deep learning models, existing works barely touch the surface of compressing DRL. In this work, we first identify that there exists an optimal model size of DRL that can maximize both the test scores and efficiency, motivating the need for task-specific DRL agents. We therefore propose an Auto-Agent-Distiller (A2D) framework, which to our best knowledge is the first neural architecture search (NAS) applied to DRL to automatically search for the optimal DRL agents for various tasks that optimize both the test scores and efficiency. Specifically, we demonstrate that vanilla NAS can easily fail in searching for the optimal agents, due to its resulting high variance in DRL training stability, and then develop a novel distillation mechanism to distill the knowledge from both the teacher agent's actor and critic to stabilize the searching process and improve the searched agents' optimality. Extensive experiments and ablation studies consistently validate our findings and the advantages and general applicability of our A2D, outperforming manually designed DRL in both the test scores and efficiency. All the codes will be released upon acceptance.

\end{abstract}

\vspace{-0.5em}
\section{Introduction}

Deep reinforcement learning (DRL)~\cite{mnih2015human} which integrates reinforcement learning (RL) and deep neural networks (DNNs) has dramatically broadened the range of complex decision-making tasks that were previously outside of the capability of machines. In particular, DRL opens up the possibility to mimic some human problem solving capabilities in high-dimensional space thanks to its capability to learn different levels of abstractions from data even with lower prior knowledge. Recent successes of DRL agents, such as beating the world-champion Go grandmaster equipped in Google's AlphaGo and beating human video game players, have triggered tremendously increased enthusiasm to develop and deploy DRL-powered intelligence into numerous real-world inference and control applications, including robotics \cite{Levine_2016, Gandhi_2017, Pinto_2017}, autonomous vehicles \cite{You_2017}, finance \cite{Deng_2017} and smart grids \cite{Francois_2017}. Many of these applications, such as autonomous vehicles, require real-time control and decision-making policies for which the DRL agents have to derive real-time policies using real-time data for dynamic systems, i.e., the policy inference must be done in real-time at the control frequency of the system, which for example is on the order of milliseconds for a recommender system \cite{Covington_2016} responding to a user request. However, real-time control and decision-making for DRL can be prohibitively challenging in many real-world applications due to DRL's integrated complex DNNs and edge devices' constrained resources, calling for DRL agent designs that favor both  test score and processing efficiency.    

In parallel, most existing DRL works adopt a fixed DNN backbone for the agent without explicitly exploring the relationship between the agents' model sizes and DRL's achieved test scores, let alone the test score and processing efficiency trade-off. While one can naturally think that increasing the model size in general will benefit the test score as commonly observed in visual classification tasks~\cite{tan2019efficientnet}, we empirically find that there exists a task-specific optimal model size for DRL's DNN agents that maximizes the achieved test score, from which further increasing the model sizes will not improve or even hurt the performance which we conjecture is due to the increased training difficulty of more complex DRL agents. This indicates that the DNN architecture for DRL agents has to balance both its model capability and imposed difficulty to DRL training, which can be very different for diverse DRL applications with different specification and task difficulty.  
Thus, task-specific DRL agent designs are highly desired to facilitate the development of optimal DRL-powered solutions that maximize both DRL's test scores and efficiency.    

To close the aforementioned gap, neural architecture search (NAS)~\cite{zoph2016neural, zoph2018learning, liu2018darts} is a promising solution as it can automate the agent design without the need for laboriously huge human efforts for each task, motivated by the recent success of AutoML~\cite{hutter2019automated}. However, directly applying NAS to design DRL agents  can easily fail due to the commonly observed vulnerability and instability of DRL training, which occurs with a high variance as discussed in ~\cite{cheng2019control, henderson2017deep, arulkumaran2017deep, recht2019tour}, making it difficult to correctly rank the sampled networks within a limited search time. Furthermore, such an instability will be further exacerbated when considering differentiable NAS (DNAS), which has the advantage of competitively low search cost and thus favors the fast development of DRL-powered solutions, because the success of DNAS requires unbiased gradient estimation with a low variance. To this end, we aim to develop a novel and effective NAS framework dedicated to DRL agent designs to promote fast development and high-quality DRL-powered intelligent solutions for numerous applications. Specifically, we make the following contributions:

\begin{itemize}
    \item We identify that there exists an optimal model size for DRL agents that maximize both the test scores and efficiency for each task, motivating task-specific DRL agent designs and calling for fast and automated DRL agent design techniques to address the growing demand for DRL-powered intelligent real-world devices. 
    
    \item We propose an Auto-Agent-Distiller (A2D) framework, which to our best knowledge is the first NAS dedicated to DRL, aiming to automatically and efficiently search for the optimal network architectures for DRL's DNN agents on different tasks. The effectiveness of A2D is attributed to a novel distillation mechanism on top of the state-of-the-art (SOTA) DRL using Actor-Critic (AC) methods, effectively stabilizing the NAS search despite the instability of DRL training. 
    
    \item Extensive experiments and ablation studies consistently validate our findings and the superiority and general applicability of our A2D, largely outperforming existing DRL in both the test scores and efficiency. We believe this work has enhanced our understanding in NAS for DRL agent design and could open up the possibility of automated and fast development of DRL-powered solutions for many real-world applications. 
    
\end{itemize}

\section{Related Works}
\subsection{Deep Reinforcement Learning}

Traditional RL algorithms can be categorized into two groups: value-based and policy-based methods. Value-based methods~\cite{watkins1992q, rummery1994line} aim to build a value function, which subsequently allows the definition of a policy; and policy-based methods~\cite{sutton2000policy} directly model the policy to maximize the expected return in either a gradient-free~\cite{gomez2005evolving, cuccu2011intrinsically} or gradient-based way~\cite{sutton2000policy}. To take advantage of both methods, AC-based RL methods~\cite{konda2000actor} try to combine both value-based and policy-based methods, in which an actor learns the feedback from a critic to trade-off both the variance introduced by policy-based methods and the bias introduced by value-based methods. Motivated by the recent success of DNNs, DRL integrates traditional RL algorithms with DNNs by approximating the optimal value function or policy using DNNs in order to scale up prior RL works to handle higher-dimensional and more complex problems. Specifically, along the value-based track, DQN~\cite{mnih2015human} first introduces DNNs to Q-Learning~\cite{watkins1992q}, and later~\cite{wang2016dueling, van2015deep, schaul2015prioritized, bellemare2017distributional, fortunato2017noisy} further improve DQN towards better value estimation and~\cite{hessel2017rainbow} combines the previous efforts to demonstrate a strong baseline; Along the policy-based track,~\cite{schulman2015trust, schulman2017proximal, liu2017stein, silver2014deterministic} develop deep policy gradient methods on top of~\cite{sutton2000policy}; And along the AC track,~\cite{mnih2016asynchronous, wang2016sample, wu2017scalable, espeholt2018impala, haarnoja2018soft} empower both the actor and critic using DNNs  and~\cite{lillicrap2015continuous, barth2018distributed, fujimoto2018addressing} further extend DRL to handle continuous control. The readers are referred to ~\cite{arulkumaran2017deep} for more details about DRL.

Despite the promising success of recent DRL methods, a majority of them focus merely on improving existing algorithms, with the DNN designs for DRL agents being under-explored. We thus aim to study the scalability of DRL with model sizes and automate the network design for DRL agents to promote fast development of powerful and efficient DRL-powered solutions for numerous applications.   

\subsection{Knowledge Transfer in DRL}
Since DRL can suffer from high sample complexity which limits its achieved performance due to the insufficient interactions with the environment, transfer learning in DRL has demonstrated to be crucial for the practical use of DRL, where external expertise knowledge is utilized to accelerate the learning process of DRL. For instance,
reward shaping~\cite{ng1999policy, wiewiora2003principled, devlin2012dynamic, brys2015policy, vecerik2017leveraging} leverages the expertise knowledge to reshape the reward distributions to steer the agent's action selection and navigation towards the expected trajectories; 
~\cite{schaal1997learning, zhang2018pretraining, kim2013learning, hester2017deep, nair2018overcoming, kang2018policy, vecerik2017leveraging} make use of external demonstrations from different sources such as human experts or near-optimal policy to achieve more efficient exploration. Motivated by the recent success of knowledge distillation~\cite{hinton2015distilling} in visual classification tasks and model compression~\cite{polino2018model}, policy distillation~\cite{rusu2015policy} has gained increased popularity in DRL, which transfers the teacher policy in a supervised manner by minimizing the divergence between the teacher policy and the student policy. More recently, ~\cite{czarnecki2019distilling, schmitt2018kickstarting} further extend this idea to enable better sample efficiency and~\cite{parisotto2015actor, teh2017distral} apply it to multi-task DRL. More information about transfer learning in DRL can be found in~\cite{zhu2020transfer}. 

Built upon the prior arts, our A2D integrates a new distillation mechanism, which not only distills the policy knowledge from the teacher actor but also the knowledge of value estimation from the teacher critic towards much improved student agent, leading to an effective NAS framework dedicated to powerful and efficient DRL agent design. 

\subsection{Neural Architecture Search}
NAS~\cite{zoph2016neural} is an exciting new field which aims to automate the network design for different tasks for achieving both competitive performances and efficiency. Specifically, RL based NAS~\cite{zoph2016neural, zoph2018learning, tan2019mnasnet, howard2019searching, tan2019efficientnet} and evolutionary algorithm based NAS~\cite{pham2018efficient, real2019regularized} explore the search space and train each sampled network candidate from scratch, suffering from prohibitive search costs. Later, DNAS~\cite{liu2018darts, wu2019fbnet, wan2020fbnetv2, cai2018proxylessnas, xie2018snas} is proposed to update the weight and architecture in a differentiable manner through supernet weight sharing, which can reduce the search time to several hours~\cite{stamoulis2019single}. Motivated by the promising performance achieved by those NAS works, 
recent works have extended NAS to more tasks such as segmentation~\cite{liu2019auto, chen2019fasterseg}, image enhancement~\cite{fu2020autogan,lee2020journey}, and language modeling~\cite{chen2020adabert}. 

The success of NAS in various tasks promise their great potential to boost DRL solutions' performance and efficiency and to enable DRL to handle large-scale real-world problems. However, to our best knowledge, existing works have not yet explored the possibility of using NAS to search for DRL agents. The main challenge is DRL's notorious training instability with a high variance as discussed in~\cite{cheng2019control, henderson2017deep, arulkumaran2017deep, recht2019tour}, which makes it difficult to correctly rank the sampled networks within a limited search time. Furthermore, although DNAS is promising in enabling fast development of powerful DRL solutions as it can best satisfy the need of fast generation and deployment of high-quality DRL agents on various tasks, the DRL's training instability often occurring with a high variance stands at odds with DNAS's requirement of accurate gradient estimation with a low variance. To this end, our A2D aims to close the aforementioned gap in a timely response to the growing need for more powerful and efficient DRL-powered solutions.

\section{Preliminaries of DRL}
\label{sec:preliminary}

In this section, we provide preliminaries of DRL during which we also set up our adopted notations. 

We assume the usual RL design, and that RL can be formulated as a Markov Decision Process (MDP) characterized using a tuple $(S, A, T, R, \gamma)$, where $S$ is the state space, A is the action space, $T(s'|s,a)$ is the transition probability of ending up in state $s'$ when executing the action $a$ in the state $s$, $R$ is the reward function, and $\gamma$ is a discount factor. The behaviour of an agent in the MDP can be formulated as a policy $ \pi({s_{t}, a_{t}}) = p(a_{t} | s_{t} )$, which defines the probability of executing the action $a$ in the state $s$. In particular, at each time step $ t $ with the corresponding state $ s_{t} \in S $, the agent performs the action $ a_{t} \in A $ sampled from the policy $ \pi({s_{t}, a_{t}})$, resulting in the state $ s_{t+1} \in S $ based on the transition probability $ T(s_{t+1}|s_t,a)$ and the reward $ r_{t} $. Accordingly, the expected cumulative reward achieved by a policy $\pi$ can be formulated as:

\begin{equation}
\label{eq:rl_cumulative_reward}
 J(\pi) = \mathbb{E}_{\pi} \left[ \sum_{t=0}^{H} \gamma^t r_{t} \right]
\end{equation}

\noindent where $ H $ is the time horizon and $ \gamma $ is the 
discount factor.
In DRL, the policy is parameterized by $ \theta_{\pi}$, i.e., the weights of the DNN agent, and the agent aims to learn an optimal policy to maximize the expected cumulative reward:

\begin{equation}
\label{eq:rl_policy_objective}
\theta_{\pi}^{*} = \underset{\theta_{\pi}}{\argmax} \,\, J(\pi(\cdot | \theta_{\pi})) 
\end{equation}

\noindent for which stochastic policy gradient methods~\cite{sutton2000policy} are widely adopted to optimize the policy and the gradient of the expected cumulative reward w.r.t. the policy parameters, i.e., $ \nabla_{\theta_{\pi}} J(\pi(\cdot | \theta_{\pi})) $, is given by:

\begin{equation}
\label{eq:rl_gradient}
\nabla_{\theta_{\pi}} J(\pi(\cdot | \theta_{\pi})) = \mathbb{E}_{\pi} \left[ \sum_{t=0}^{H} \delta_{t} \nabla_{\theta_{\pi}} \log(\pi(a_t, s_t | \theta_{\pi})) \right]
\end{equation}

\noindent where there are different ways to design $\delta_{t}$ and 
we adopt the temporal difference error (td-error) following~\cite{sutton1988learning} for reducing the variance of the policy gradients, i.e.:

\begin{equation}
\label{eq:delta_t}
\delta_{t} = r_{t} + \gamma V_{\pi}(s_{t+1} | \theta_{v}) - V_{\pi}(s_{t} | \theta_{v})
\end{equation}

\noindent where $ V_{\pi}(s) = \mathbb{E}_{\pi} \left[ \sum_{t=0}^{H} \gamma^{t} r_{t} | s_0 = s \right] $ is the value function which estimates the expected cumulative reward of the policy $\pi$ starting from the initial state $s$.

\begin{figure*}[!t]
\begin{center}
   \includegraphics[width=1.0\linewidth]{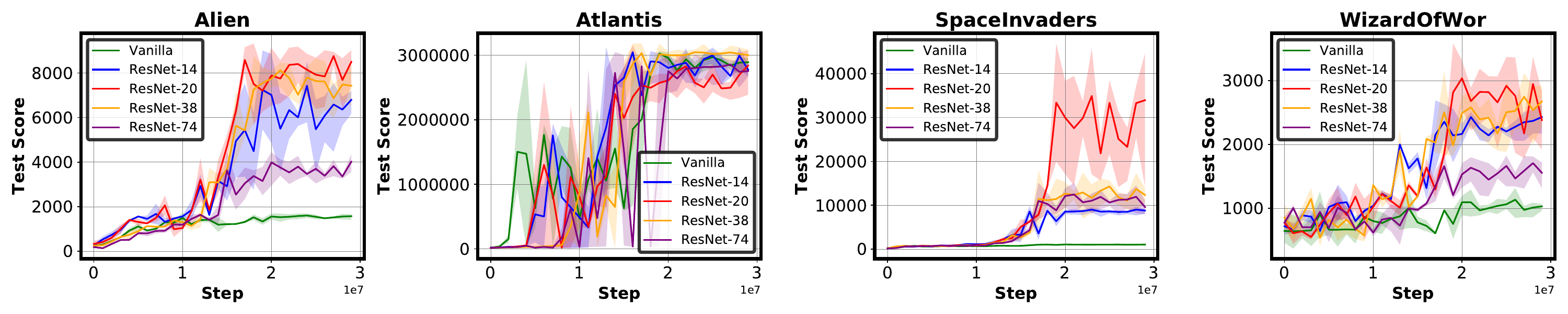}
\end{center}
  \vspace{-1.5em}
   \caption{Test scores averaged over 30 episodes during the training of five models on four Atari games~\cite{bellemare2013arcade}.}
   \vspace{-1.5em}
\label{fig:scalability}
\end{figure*}

Since the value function is also unknown, the AC-based methods~\cite{konda2000actor} approximate the value function (i.e., the critic) with the learnable parameter $ \theta_{v} $ which can be modeled using the weights of a DNN, and the learning objective of $ \theta_{v} $ is to minimize the td-error of the estimated value between consecutive states, as formulated below: 

\begin{equation}
\label{eq:rl_value_objective}
\theta_{v}^{*}  = \underset{\theta_{v}}{\argmin} \,\, \mathbb{E}_{\pi} \left[ \sum_{t=0}^{H} \frac{1}{2} \left( r_{t} + \gamma V_{\pi}(s_{t+1} | \theta_{v}) - V_{\pi}(s | \theta_{v}) \right)^2 \right] \\ 
\end{equation}

\noindent Therefore, the actor and critic parameterized by $\theta_{\pi}$ and $ \theta_{v}$, respectively can be updated in an iterative way to guide the agent towards an optimal policy.

\section{Motivating Findings and Proposed Methods}

In this section, we first present two motivating observations that motivate our proposed A2D framework based on experiments and discussions, i.e., the necessity of task-specific DRL agent designs in Sec.~\ref{sec:scalability} and the failure of vanilla NAS in DRL in Sec.~\ref{sec:failure}, and then introduce our proposed AC-based distillation mechanism and the A2D framework in Sec.~\ref{sec:distillation} and Sec.~\ref{sec:a2d}, respectively.

\subsection{Finding 1: Scalability of DRL with Model Sizes}
\label{sec:scalability}

In this subsection, we provide experiments and discussions to show the need for task-specific DRL agent designs by studying the scalability of DRL with model sizes. 

Most of existing DRL works adopt a fixed network backbone for the agents without explicitly studying the effect of model sizes to the final performance, leaving the scalability of DRL with model sizes unexplored and limiting the potential performance of DRL. To study this effect, we evaluate one of the most representative DRL, the AC-based DRL~\cite{konda2000actor} method, when using various network backbones.

\textbf{Agent Backbone Design.} We follow the widely adopted network backbone design in existing AC-based DRL methods~\cite{espeholt2018impala, mnih2016asynchronous}, which uses a heavy feature extractor and two light-headers (implemented using two fully-connected layers) to design the actor and critic, respectively. Motivated by the success of SOTA DNN models, ResNet~\cite{he2016deep} series, as also demonstrated in recent DRL works~\cite{espeholt2018impala}, we adopt ResNets as the feature extractor, using the standard ResNet designed for the CIFAR-10 dataset~\cite{he2016identity}, and scale up the model sizes by increasing its depth, during which we slightly modify the strides as well as the first and last layer of the networks to adapt the networks to the target DRL tasks as detailed in Appendix.~\ref{appendix:network}.

\begin{figure*}[!t]
\begin{center}
   \includegraphics[width=1.0\linewidth]{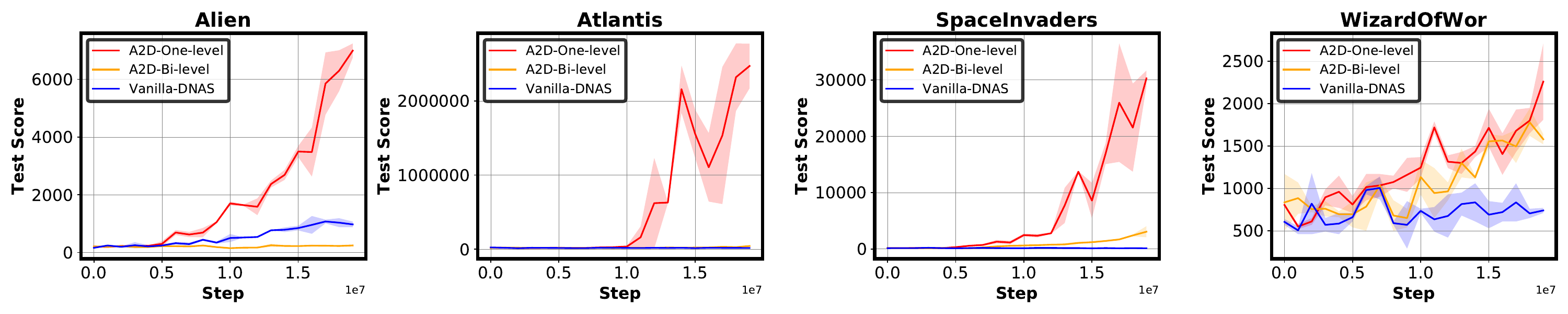}
\end{center}
  \vspace{-1.5em}
   \caption{Test score evolution during the search processes of three different search schemes on four Atari games~\cite{bellemare2013arcade}, where Vanilla-DNAS denotes directly applying NAS without distillation, and A2D-One-level and A2D-Bi-level search under the guidance of the distillation loss using one-level and bi-level optimization, respectively.}
   \vspace{-1.5em}
\label{fig:nas}
\end{figure*}

\textbf{Evaluation Settings.} Note that we use the same training and test hyper-parameters settings for all the models on all the tasks in this paper. \underline{Models and tasks:} we evaluate the performance of the AC-based DRL when its feature extractor backbone adopts five different networks with different model sizes, including the original small network in DQN~\cite{mnih2015human} (termed as the Vanilla), ResNet-14, ResNet-20, ResNet-38, and ResNet-74, on Atari 2600 games based on the Arcade Learning Environment~\cite{bellemare2013arcade}. \underline{Training settings:} we train a DRL agent on each task for 3e7 steps with a discount factor ($\gamma$ in Eq.~\ref{eq:rl_cumulative_reward}) of 0.99 and a rollout length of 5. We use the RMSProp optimizer as~\cite{mnih2015human} with an initial learning of 1e-3 which keeps constant in the first 1e7 steps and then linearly decays to 1e-4. We adopt the standard loss for the actor and critic as in Eq.~\ref{eq:rl_policy_objective} and Eq.~\ref{eq:rl_value_objective} as well as an entropy loss on top of the actor to encourage exploration as in~\cite{haarnoja2018soft}.
\underline{Test settings:} the reported test score is averaged on 30 episodes with null-op starts following~\cite{mnih2015human}.

\begin{table}[!t]
\vspace{0.2em}
\caption{The best test scores achieved by different models on 15 Atari games. More results can be found in Appendix.~\ref{appendix:scalability}.}
  \resizebox{1.0\linewidth}{!}
  {    
\begin{tabular}{c|ccccc}
\hline
& Vanilla & ResNet-14 & ResNet-20 & ResNet-38 & ResNet-74 \\ \hline \hline
Alien & 1724 & 9007 & \textbf{9323} & 8829 & 4456 \\ 
Amidar & 721.8 & 638.6 & \textbf{845.2} & 138.3 & 493.9 \\
Assault & 10164 & 14470 & \textbf{17314} & 12406.5 & 9849 \\
Asterix & 4850 & 708500 & \textbf{856800} & 756120 & 539060 \\ 
Asteroids & 2095 & 5690 & \textbf{5744} & 1947 & 4792 \\
BankHeist & 1152 & \textbf{1288} & 1284 & 981 & 1163 \\  
BattleZone & 7600 & 5800 & 13100 & \textbf{13300} & 4100 \\ 
BeamRider & 5530 & 23984 & 25961 & 29498 & \textbf{30048} \\ 
Boxing & 4.2 & \textbf{100} & \textbf{100} & 99.3 & 87.1 \\
ChopperCommand & 1320 & 11170 & \textbf{14910} & 4370 & 8240 \\ 
DemonAttack & 318349 & 481818 & 484382 & \textbf{494569} & 448450 \\ 
Pong & -19.9 & \textbf{21} & 20.9 & 20.9 & 20.8 \\ 
SpaceInvaders & 1171 & 9848 & \textbf{46870} & 17962 & 15111 \\
Tennis & -23.7 & 13.8 & 11.5 & \textbf{19.6} & 19.3 \\ 
WizardOfWor & 1320 & 2690 & \textbf{3580} & 3160 & 1850 \\ \hline
\end{tabular}
    }
  \label{tab:scalability}
  \vspace{-1.5em}
\end{table}

\textbf{Observations and Analysis.} We visualize the test score evolution during the training process on various Atari games, when adopting different networks for the DRL agents as shown in Fig.~\ref{fig:scalability}, and show the highest achieved test scores in Tab.~\ref{tab:scalability}. We can observe that \underline{(1)} larger model sizes with a higher network capability generally benefit the achieved test scores especially on more complex games (e.g., BeamRider in Tab.~\ref{tab:scalability}), as larger models can achieve higher test scores under the same training time steps in most of the games as compared with the smaller vanilla-network and ResNet-14, and \underline{(2)} there always exists a task-specific optimal model size where larger models cannot further improve or even degrade the test score, which we conjecture is due to the increased training difficulty associated with larger DRL agents. For instance, the vanilla network can achieve decent test scores in the Atlantis game (see Fig.~\ref{fig:scalability}), while ResNet-38 can only marginally improve the score with 13.7$\times$ higher FLOPs (floating-point operations). \underline{In addition}, ResNet-74 performs worse than ResNet-20/38 in most of these experiments, due to its increased training difficulty within the limited training steps. We recognize that more training steps and well-tuned training hyper-parameters may help the case with ResNet-74 converge to a better optima, however the resulting inefficiency in both training and inference as compared to more efficient agents will limit the practical usability of such solutions.

Experiments with different models on more tasks are provided in Appendix.~\ref{appendix:scalability}.

\textbf{Extracted Insights.} The above observations indicate that (1) the DNN architecture is a critical design factor in DRL despite their being under-explored in existing works, and (2) it is highly desired to design task-specific agents which optimally balance the test score and processing efficiency for different tasks. Meanwhile, manually designing dedicated agents for different tasks is time/labor-consuming, and thus not practical for addressing the growing demand for the fast development and deployment of DRL-powered intelligent solutions for numerous applications, motivating the introduction of NAS into the DRL domain to automate the design of optimal DRL agents.

\subsection{Finding 2: Vanilla DNAS Fails in DRL}
\label{sec:failure}

In this subsection, we present experiments and discussions to show the challenges and failures of directly applying vanilla DNAS to design DRL agents. 

\label{sec:failure}
\textbf{Challenges of Applying DNAS to DRL.} 
We mainly consider DNAS~\cite{liu2018darts} since its low search time and cost make it better align with our target goal of fast development and deployment of powerful and efficient DRL agents for different tasks, as compared with other NAS methods that require higher search costs and may limit the number of interactions with the environments. Different from other tasks such as visual classification that are facilitated with consistent supervision information, the training of DRL is often unstable due to the high variance in the gradient estimation as discussed in~\cite{cheng2019control, henderson2017deep, arulkumaran2017deep, recht2019tour}, making it difficult to correctly rank the sampled networks within a limited search time. This challenge can be further exacerbated when applying DNAS to DRL, as effective DNAS requires unbiased and stable (with low variance) gradient estimation~\cite{bi2019stabilizing,he2020milenas} to guarantee the convergence while such a requirement is often not feasible in a DRL environment, especially in real-world applications.

\textbf{Evaluation Settings of Applying Vanilla DNAS.} Here we use the same search space and search settings as those for evaluating the proposed A2D framework described in Sec.~\ref{sec:a2d} where more details can be found. In general, to evaluate the feasibility of directly applying vanilla DNAS to DRL, we adopt the search strategy in~\cite{xie2018snas} where a Gumbel Softmax~\cite{jang2016categorical} function is utilized to sample candidate operators from the search space and a one-level optimization to update the weight and architecture parameters with unbiased gradient estimation~\cite{he2020milenas}.

\textbf{Observations and Analysis.}
Fig.~\ref{fig:nas} (see the blue line) visualizes the test score evolution during the search process. We can observe that the supernet can hardly converge with the consistently low rewards close to those of using random exploration among all the tasks, indicating that the supernet cannot serve as a reliable proxy to evaluate the performance of the derived networks. These observations validate our conjecture that the high variance in DRL training naturally forbids the practicality of vanilla DNAS in DRL.

\subsection{The Proposed Distillation Mechanism}
\label{sec:distillation}
As discussed in Sec.~\ref{sec:failure}, the key challenge of applying DNAS in DRL is the search instability caused by the high variance of the gradient estimates of DRL. Inspired by~\cite{rusu2015policy, parisotto2015actor} which show that the distillation from a teacher agent can effectively reduce the variance of gradient estimates and stabilize the training process of the student agent, we propose a distillation mechanism based on the AC-based DRL method, aiming to stabilize the NAS process and improve the optimality of the resulting DRL agents.

\textbf{Motivation.} Policy distillation~\cite{rusu2015policy} is one of the most natural choices thanks to both its simplicity across different tasks and its differentiable property that aligns with our goal of applying DNAS to DRL. However, vanilla policy distillation merely distills the policy without considering the value function which can play a critical role in both assisting the policy updates and reducing the variance of vanilla policy gradients. Furthermore, from the network design perspective, the value function can be viewed as a high-level semantic information on top of the extracted features which can thus provide important signals to guide the training of the feature extractor network, similar to the effect of the perceptual loss in GANs~\cite{johnson2016perceptual}. We thus hypothesize that further distilling the value function from the teacher agent can better improve the training stability and the convergence.

\begin{table*}[t]
\caption{Benchmarking different distillation mechanisms for training the vanilla network and ResNet-14 with the guidance from a pretrained ResNet-20 on six Atari games~\cite{bellemare2013arcade}, where row No.4 and No.8 are the proposed method. More results can be found in Appendix.~\ref{appendix:distillation}.}
\label{table:distillation}
\begin{threeparttable}
\centering
\resizebox{1\textwidth}{!}{
{
\begin{tabular}{c|cccc||cccccc} 
\toprule
 \multirow{2}{*}{\textbf{Model}} & \multirow{2}{*}{\textbf{No.}} &  \textbf{Distill} & \textbf{Reuse} & \textbf{Distill the Critic} &  \multirow{2}{*}{\textbf{Alien}} & \textbf{Space-} & \textbf{Wizard-} & \multirow{2}{*}{\textbf{Asterix}} & \textbf{Battle-} & \textbf{Beam-} \\ 
&  & \textbf{the Actor} & \textbf{Teacher Critic} & \textbf{with MSE Loss} &   & \textbf{-Invaders}  &  \textbf{-OfWor}  & & \textbf{-Zone}  & \textbf{-Rider} \\ 
\midrule
\multirow{4}{*}{\textbf{Vanilla}} & 1 & & & & 1724 & 1171 & 1320 & 4850 & 7600 & 5530 \\
&2  & \checkmark & & & 3096 & 26821 & 5560 & 59020 & 12700 & 14417 \\
&3  & \checkmark & \checkmark & & 249 & 30124 & \textbf{6310} & 37780 & 14200 & 15432 \\
& \textbf{4 (Proposed)}  &\checkmark & & \checkmark & \textbf{3419} & \textbf{39274} & 5960 & \textbf{64510} & \textbf{14500} & \textbf{17806} \\
\midrule
\multirow{4}{*}{\textbf{ResNet-14}} & 5 & & & & 9007 & 9848 & 2690 & 708500 & 5800 & 23984 \\
& 6& \checkmark & & & 14682 & 76246 & \textbf{6300} & 749870 & 16300 & 38311 \\
& 7& \checkmark& \checkmark & & 12674 & 79287 & 4640 & 695550 & 16400 & 37350 \\
& \textbf{8 (Proposed)} & \checkmark & & \checkmark & \textbf{15723} & \textbf{111189} & 5450 & \textbf{849400} & \textbf{18200} & \textbf{42365} \\
 \bottomrule 
\end{tabular}
}
}
\end{threeparttable}
\vspace{-1em}
\end{table*}

\textbf{Design of Our Distillation Mechanism.} We propose an AC-based distillation mechanism, where we first train an AC agent using a large network backbone and then guide the training of the student agent by distilling from both the actor and critic of the teacher agent. Specifically, the two distillation losses are formulated as:

\begin{equation}
\label{eq:loss_actor}
L_{actor}^{distill} = \mathbb{E}_{\pi} \left[ \sum_{t=0}^{H} \pi(a_t, s_t | \theta_{\pi}^{tea}) \,\,
log \frac{\pi(a_t, s_t | \theta_{\pi}^{tea})}{ \pi(a_t, s_t | \theta_{\pi}^{stu})} \right]
\end{equation}

\begin{equation}
\label{eq:loss_critic}
L_{critic}^{distill}  =  \mathbb{E}_{\pi} \left[ \sum_{t=0}^{H} \frac{1}{2} \left( V_{\pi}(s_{t} | \theta_{v}^{stu}) - V_{\pi}(s_{t} | \theta_{v}^{tea}) \right)^2  \right]
\end{equation}

\noindent where $\pi(a_t, s_t | \theta_{\pi}^{tea})$ and $\pi(a_t, s_t | \theta_{\pi}^{stu}) $ are the teacher and student actor, respectively, and $V_{\pi}(s_{t} | \theta_{v}^{tea})$ and $V_{\pi}(s_{t} | \theta_{v}^{stu})$ are the teacher and student critic, respectively. Here we adopt KL divergence to distill the knowledge from the teacher actor following~\cite{rusu2015policy} and the MSE loss as a soft constraint to enforce the student critic to incorporate the estimated value of the teacher critic. Therefor, the training objective during both search and training is:

\begin{equation}
\begin{split}
\label{eq:loss_total}
L_{total}  = \,\, & L_{policy} + L_{value} + \alpha_{1}L_{entropy} \\ 
 + \,\, & \alpha_{2} L_{actor}^{distill} + \alpha_{3}L_{critic}^{distill}
\end{split}
\end{equation}

\noindent where $\alpha_1$, $\alpha_2$, and $\alpha_3$ are the weighted coefficients. 
As in Sec.~\ref{sec:preliminary}, here $L_{policy} = \mathbb{E}_{\pi} \left[ - \sum_{t=0}^{H} \delta_{t} \log(\pi(a_t, s_t | \theta_{\pi}^{stu})) \right]$ is the policy gradient loss as in~\cite{sutton2000policy}, $L_{value} = \mathbb{E}_{\pi} \left[ \sum_{t=0}^{H} \frac{1}{2} \left( r_{t} + \gamma V_{\pi}(s_{t+1} | \theta_{v}^{stu}) - V_{\pi}(s_t | \theta_{v}^{stu}) \right)^2 \right]$ is the value loss based on the td-error~\cite{sutton1988learning}, and $L_{entropy} = \mathbb{E}_{\pi} \left[ \sum_{t=0}^{H} \pi(a_t, s_t | \theta_{\pi}^{stu}) \,\, log(\pi(a_t, s_t | \theta_{\pi}^{stu})) \right] $  is the entropy loss on top of the policy to encourage exploration as in~\cite{haarnoja2018soft}.




\textbf{Discussions about the Distillation Design.} Another choice for distilling the knowledge from the teacher critic is to directly apply the estimated value of the teacher critic as that of the student critic, i.e., without training a new student critic. However, this may lead to an overestimation of the value if the teacher critic is inaccurate as analyzed in~\cite{van2015deep, thrun1993issues, hasselt2010double, van2011insights}, the resulting approximation error of which will further accumulate in the student agent. Therefore, instead of completely inheriting the teacher critic, we apply an MSE loss between the estimated value of the student and teacher critics to distill the knowledge in a soft manner, mitigating the potential of overestimation as validated in our evaluation experiments.

\textbf{Evaluation Settings.} We use the same training settings as in Sec.~\ref{sec:scalability} except that we incorporate the distillation loss as in Eq.~\ref{eq:loss_total}. In particular, we train a ResNet-20 model as the teacher agent for all the experiments and $\alpha_1$, $\alpha_2$, and $\alpha_3$ in Eq.~\ref{eq:loss_total} are set to be 1e-2, 1e-1, and 1e-3, respectively, in all our settings. We benchmark the proposed distillation mechanism with three baselines: (1) the original training scheme without distillation, (2) training with the distillation only from the teacher actor, i.e., $\alpha_3$ is zero, as in~\cite{rusu2015policy}, and (3) training with the distillation from the teacher actor and directly reusing the estimated value from the teacher critic without training a new critic.

\textbf{Observations and Analysis.}
We evaluate the proposed distillation mechanism and the three baselines by applying them to the vanilla network and ResNet-14 evaluated on six Atarix games as shown in Tab.~\ref{table:distillation}. We can observe that: (1) different distillation strategies generally improve the test scores compared with the ones without distillation, which is consistent with~\cite{rusu2015policy}; (2) Among the three distillation strategies, our proposed distillation mechanism consistently outperforms the other two strategies in achieving higher test scores on most of tasks, and (3) reusing the teacher critic combined with the distillation from the teacher actor (i.e., row No.3 and No.7 in Tab.~\ref{table:distillation}) suffers from negative effects in six out of the 12 cases compared with the case of only distilling the actor, verifying that our hypothesized overestimation problem can occur if merely inheriting the teacher critic without training a student critic.

More benchmark experiments over the SOTA policy distillation~\cite{rusu2015policy} are provided in Appendix.~\ref{appendix:distillation}.

\subsection{The Proposed A2D Framework}
\label{sec:a2d}

In this subsection, we describe our A2D framework, which to our best knowledge is the first NAS framework for DRL and makes use of the distillation mechanism presented in Sec. \ref{sec:distillation}. Specifically, A2D integrates the knowledge distilled from a teacher agent to stabilize the search process of NAS and adopts DNAS considering its advantageous search efficiency following~\cite{liu2018darts, wu2019fbnet}, which can be formulated as a one-level optimization problem: 

\begin{align} 
    \begin{split}
    & \min \limits_{\theta_{\pi}, \theta_{v}, \alpha} \,\, L_{total}(\theta_{\pi}, \theta_{v}, \alpha)+\lambda L_{cost}(\alpha) \label{eq:a2d_objective} 
    \end{split}
\end{align}

\noindent where $L_{total}$ is the distillation-based loss (see Eq.~\ref{eq:loss_total}), $L_{cost}$ is an efficiency loss (e.g., the total number of floating-point operations (FLOPs)) weighted by the coefficient $\lambda$ and determined by the searched network architecture, and $\alpha$ is the architecture parameter which stores the probability of each candidate operator. 

\textbf{Optimization Method of A2D.} As shown in Eq.~\ref{eq:a2d_objective}, We adopt a one-level optimization in A2D, i.e., update the weight and architecture parameters in the same iteration as in~\cite{xie2018snas}, instead of using a bi-level optimization as in~\cite{liu2018darts}. Our consideration is that the bi-level optimization induces biased gradient estimation due to the approximation resulting from its one-step stochastic gradient descent~\cite{ he2020milenas, bi2019stabilizing}, which may harm the search stability especially in the DRL context. In contrast, the one-level optimization has unbiased gradient estimation~\cite{he2020milenas} and its potential over-fitting problem can be alleviated by the randomness introduced by the Gumbel Softmax function~\cite{jang2016categorical} as widely adopted in~\cite{xie2018snas, hu2020dsnas, bi2020gold}. We will benchmark these two optimization methods in the experiments.

\textbf{The NAS Sampling Method.} During the DNAS search, each layer in the supernet is formulated as a weighted sum of the outputs from all the $N$ candidate operators, i.e.,

\begin{align} 
    \begin{split}  \label{eq:supernet_layer}
    A^{l+1} &= \sum_{i=1}^N  GS(\alpha_i^l) O_i^l(A^l)
    \end{split} \\
    \begin{split}  \label{eq:gumbel_softmax}
 \text{where} \,\, GS(\alpha_i^l) &=\frac{\exp{[(log(\alpha_i^l)+g_i^l)/\tau]}}{\sum_i\exp{[(log(\alpha_i^l)+g_i^l)/\tau]}}
    \end{split} 
\end{align}

\noindent where $A^l$ and $A^{l+1}$ are the output activations of two consecutive layers, respectively; $O_i^{l}$ is the $i$-th candidate operator in the $l$-th layer whose probability is controlled by $\alpha_i^l$; $GS$ is the Gumbel Softmax function~\cite{jang2016categorical} which relaxes the discrete architecture distribution to be continuous and differentiable using the reparameterization trick; $g_i^l$ is a random variable sampled from the Gumbel distribution; and $\tau$ is a temperature parameter which is annealing during the search process to enforce $GS(\alpha_i^l)$ to be close to an one-hot vector at the end of search for narrowing the gap between the supernet and the derived network.

The search algorithm of A2D is summarized in Alg.~\ref{alg:a2d}.

\begin{algorithm}[t!]
\caption{Auto-Agent Distiller (A2D): Searching for Efficient DRL Agents}
\begin{algorithmic}
\small
\Require the total steps $T_{max}$, the rollout length $L$, the discount factor $\gamma$, the teacher actor and critic $\pi(\cdot | \theta_{\pi}^{tea})$ and $V_{\pi}(\cdot | \theta_{v}^{tea})$, the learning rates of supernet weights and architecture parameters $\eta_1$ and $\eta_2$, respectively, the weight of efficiency loss $\lambda$

\State Initialize $\pi(\cdot | \theta_{\pi}^{stu})$, $V_{\pi}(\cdot | \theta_{v}^{stu})$ and the architecture $\alpha$
\State Initialize the step counter $t\gets 1$
\Repeat
\State $t_{start} = t$
\State Get state $s_t$
\Repeat
\State Perform $a_t \sim \pi(a_t, s_t | \theta_{\pi}^{stu})$ based on Eq.~\ref{eq:supernet_layer} and Eq.~\ref{eq:gumbel_softmax}
\State Receive reward $r_t$ and new state $s_{t+1}$
\State $t \gets t + 1$
\Until terminal $s_t$ \textbf{or} $t-t_{start} == L$
\For {$i \in \{t_{start},\ldots,t-1\}$}
\State $ \delta_{t} = r_{t} + \gamma V_{\pi}(s_{t+1} | \theta_{v}^{stu}) - V_{\pi}(s_t | \theta_{v}^{stu})$
\State Calculate $L_{total}$ in Eq.~\ref{eq:loss_total} based on $\delta_{t}$, $\pi(a_t,s_t | \theta_{\pi}^{tea})$, and $V_{\pi}(s_t | \theta_{v}^{tea})$
\State Update $\theta_{\pi}^{stu}$: $\theta_{\pi}^{stu} \gets \theta_{\pi}^{stu} - \eta_1 \nabla_{\theta_{\pi}^{stu}} L_{total}$
\State Update $\theta_{v}^{stu}$: $\theta_{v}^{stu} \gets \theta_{v}^{stu} - \eta_1 \nabla_{\theta_{v}^{stu}} L_{total}$
\State Update $\alpha$: $\alpha \gets \alpha - \eta_2 \nabla_{\alpha} (L_{total}+\lambda L_{cost})$
\EndFor
\Until $t > T_{max}$
\State Derive the final agent with the highest $\alpha$

\noindent\Return the final agent

\end{algorithmic}
\label{alg:a2d}
\end{algorithm}

\textbf{Evaluation settings.} \underline{Search space:} the supernet structure follows the network design in Sec.~\ref{sec:scalability} with a searchable feature extractor and two light-weight headers served as the actor and critic, respectively. Specifically, we adopt a sequential block-based supernet structure as ~\cite{wu2019fbnet} for facilitating hardware efficiency, which includes 14 searchable cells with each containing 9 candidate operators, leading to a search space of $9^{14}$ choices. More details about the supernet structure are provided in Appendix.~\ref{appendix:supernet}. \underline{Search settings:} we update the architecture parameters using an Adam optimizer with a momentum of 0.9 and a fixed learning rate of 1e-3. The initial temperature $\tau$ in Eq.~\ref{eq:gumbel_softmax} is set to 5 and decayed by 0.98 every 1e5 steps to ensure the final temperature to be close to 0.1. The training settings of the weights and the objective function are the same as in Sec.~\ref{sec:distillation} except that we only search for 2e7 steps for an acceptable search time. \underline{Derive the final agent:} we select the operators with the highest probability indicated by $\alpha$ to derive the searched agent, and then train it from scratch with the proposed distillation mechanism following the training setting in Sec.~\ref{sec:distillation} under the guidance of a pretrain ResNet-20 agent.

\begin{figure*}[!t]
\begin{center}
   \includegraphics[width=0.95\linewidth]{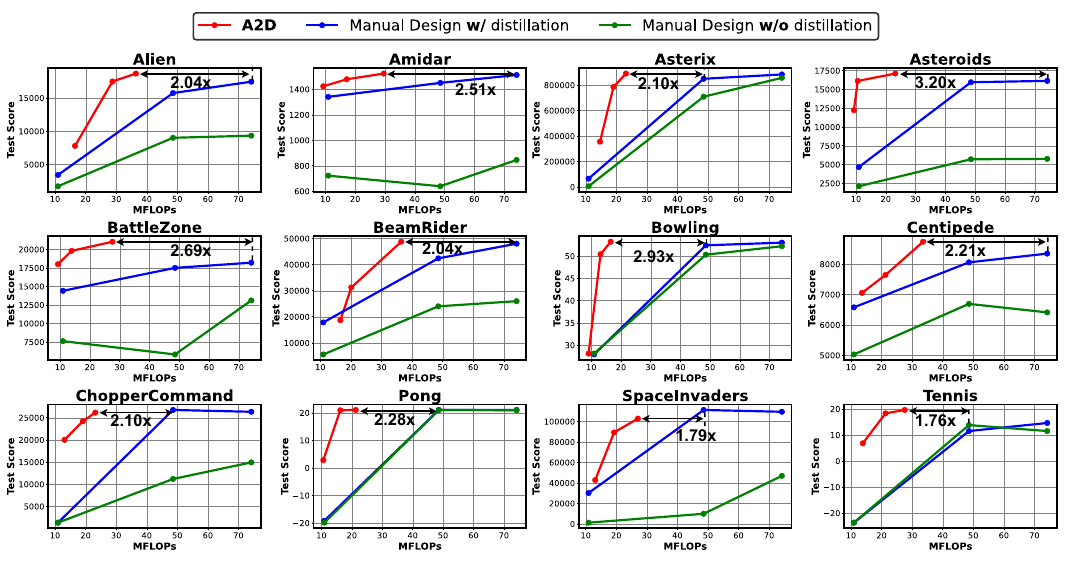}
\end{center}
\vspace{-1.5em}
\caption{Test score and efficiency trade-offs achieved by our A2D and the manually designed DRL on 12 Atari games~\cite{bellemare2013arcade}, where the three manual designs use the vanilla network, ResNet-14, and ResNet-20 introduced in Sec.~\ref{sec:scalability}.}
\vspace{-1em}
\label{fig:trade-off}
\end{figure*}

\textbf{Observations and Analysis.} 
\underline{Test score evolution} during search: we visualize the test score evolution searched using both the one-level and bi-level optimization, respectively, with both under the guidance of the distillation loss, as shown in Fig.~\ref{fig:nas}. We can observe that the test scores remain low when searching with the bi-level optimization, validating our mentioned hypothesis that the supernet cannot serve as an accurate proxy to indicate the performance of the sampled subnetworks. In contrast, searching with the one-level optimization leads to a consistent improvement in the test scores during search, demonstrating the first framework that successfully makes NAS possible in DRL.

\underline{Test score vs. efficiency trade-offs} on 12 Atari games: Fig.~\ref{fig:trade-off} shows the test score and efficiency trade-offs of both the networks searched by the proposed A2D and the manual designs trained w/ and w/o the proposed distillation mechanism, the latter of which includes the vanilla network, ResNet-14, and ResNet-20 as introduced in Sec.~\ref{sec:scalability}.
We can observe that (1) our A2D framework achieves comparable test scores with the best baseline at the cost of 1.76$\times$ $\sim$ 3.20$\times$ fewer FLOPs, and (2) the proposed A2D can derive DRL agents with different model sizes to automatically adapt to the task difficulty, e.g., fewer than 20 MFLOPs for the Bowling game and near 40 MFLOPs for the Alien game as shown in Fig.~\ref{fig:trade-off}, verifying that A2D is capable to identify the optimal agent model sizes depending on the target tasks to maximize both the test score and processing efficiency. Although the vanilla network is also promising in terms of the required FLOPS, it requires a larger parameter number, e.g., 5.83$\times$ compared with ResNet-20, and more than 10$\times$ larger than the searched networks of A2D. 



\textbf{Discussions.} Our A2D framework is to study the practicality and possibility of applying DNAS to search for DRL agents. While the above experiments consistently validate its effectiveness over manually expert-designed DRL agents,
we recognize that A2D is merely one very first step towards automated design of DRL agents. Many further explorations can be performed on top of A2D, including 
the exploration of (1) larger and more complex search spaces, (2) different DRL recipes (e.g., on-policy/off-policy), (3) transferability to other DRL algorithms, and (4) the extension to real-world applications (e.g., autonomous driving), which we leave for our future work.

 \vspace{-0.7em}
\section{Conclusion}
 \vspace{-0.3em}
The recent breakthroughs of DRL agents have motivated 
a growing demand for deploying DRL agents into intelligent devices for real-time control while the prohibitive computational cost of DRL stands at the odds with the limited on-device resources. 
To close this gap, we first identify that there exists a task-specific optimal model size that maximizes both the test scores and efficiency, and then propose the A2D framework, which to our best knowledge is the first NAS applied to DRL to automatically search for the optimal DRL agents for various tasks based on a novel distillation mechanism to stabilize the notorious searching process and improve optimality.
Extensive experiments and ablation studies consistently validate the superiority of the proposed A2D framework over manually expert-designed DRL in optimizing the test scores and processing efficiency. This work enhances our understanding in NAS for DRL and can open up the possibility of automated and fast development of DRL-powered solutions for real-world applications.


{\small
\bibliographystyle{cvpr2021}
\bibliography{ref}

\begin{thebibliography}{10}\itemsep=-1pt

\bibitem{arulkumaran2017deep}
Kai Arulkumaran, Marc~Peter Deisenroth, Miles Brundage, and Anil~Anthony
  Bharath.
\newblock Deep reinforcement learning: A brief survey.
\newblock {\em IEEE Signal Processing Magazine}, 34(6):26--38, 2017.

\bibitem{barth2018distributed}
Gabriel Barth-Maron, Matthew~W Hoffman, David Budden, Will Dabney, Dan Horgan,
  Dhruva Tb, Alistair Muldal, Nicolas Heess, and Timothy Lillicrap.
\newblock Distributed distributional deterministic policy gradients.
\newblock {\em arXiv preprint arXiv:1804.08617}, 2018.

\bibitem{bellemare2017distributional}
Marc~G Bellemare, Will Dabney, and R{\'e}mi Munos.
\newblock A distributional perspective on reinforcement learning.
\newblock {\em arXiv preprint arXiv:1707.06887}, 2017.

\bibitem{bellemare2013arcade}
Marc~G Bellemare, Yavar Naddaf, Joel Veness, and Michael Bowling.
\newblock The arcade learning environment: An evaluation platform for general
  agents.
\newblock {\em Journal of Artificial Intelligence Research}, 47:253--279, 2013.

\bibitem{bi2019stabilizing}
Kaifeng Bi, Changping Hu, Lingxi Xie, Xin Chen, Longhui Wei, and Qi Tian.
\newblock Stabilizing darts with amended gradient estimation on architectural
  parameters.
\newblock {\em arXiv preprint arXiv:1910.11831}, 2019.

\bibitem{bi2020gold}
Kaifeng Bi, Lingxi Xie, Xin Chen, Longhui Wei, and Qi Tian.
\newblock Gold-nas: Gradual, one-level, differentiable.
\newblock {\em arXiv preprint arXiv:2007.03331}, 2020.

\bibitem{brys2015policy}
Tim Brys, Anna Harutyunyan, Matthew~E Taylor, and Ann Now{\'e}.
\newblock Policy transfer using reward shaping.
\newblock In {\em Proceedings of the 2015 International Conference on
  Autonomous Agents and Multiagent Systems}, pages 181--188, 2015.

\bibitem{cai2018proxylessnas}
Han Cai, Ligeng Zhu, and Song Han.
\newblock Proxylessnas: Direct neural architecture search on target task and
  hardware.
\newblock {\em arXiv preprint arXiv:1812.00332}, 2018.

\bibitem{chen2020adabert}
Daoyuan Chen, Yaliang Li, Minghui Qiu, Zhen Wang, Bofang Li, Bolin Ding, Hongbo
  Deng, Jun Huang, Wei Lin, and Jingren Zhou.
\newblock Adabert: Task-adaptive bert compression with differentiable neural
  architecture search.
\newblock {\em arXiv preprint arXiv:2001.04246}, 2020.

\bibitem{chen2019fasterseg}
Wuyang Chen, Xinyu Gong, Xianming Liu, Qian Zhang, Yuan Li, and Zhangyang Wang.
\newblock Fasterseg: Searching for faster real-time semantic segmentation.
\newblock {\em arXiv preprint arXiv:1912.10917}, 2019.

\bibitem{cheng2019control}
Richard Cheng, Abhinav Verma, Gabor Orosz, Swarat Chaudhuri, Yisong Yue, and
  Joel~W Burdick.
\newblock Control regularization for reduced variance reinforcement learning.
\newblock {\em arXiv preprint arXiv:1905.05380}, 2019.

\bibitem{Covington_2016}
Paul Covington, Jay Adams, and Emre Sargin.
\newblock Deep neural networks for youtube recommendations.
\newblock In {\em Proceedings of the 10th ACM Conference on Recommender
  Systems}, New York, NY, USA, 2016.

\bibitem{cuccu2011intrinsically}
Giuseppe Cuccu, Matthew Luciw, J{\"u}rgen Schmidhuber, and Faustino Gomez.
\newblock Intrinsically motivated neuroevolution for vision-based reinforcement
  learning.
\newblock In {\em 2011 IEEE International Conference on Development and
  Learning (ICDL)}, volume~2, pages 1--7. IEEE, 2011.

\bibitem{czarnecki2019distilling}
Wojciech~Marian Czarnecki, Razvan Pascanu, Simon Osindero, Siddhant~M
  Jayakumar, Grzegorz Swirszcz, and Max Jaderberg.
\newblock Distilling policy distillation.
\newblock {\em arXiv preprint arXiv:1902.02186}, 2019.

\bibitem{Deng_2017}
Y. {Deng}, F. {Bao}, Y. {Kong}, Z. {Ren}, and Q. {Dai}.
\newblock Deep direct reinforcement learning for financial signal
  representation and trading.
\newblock {\em IEEE Transactions on Neural Networks and Learning Systems},
  28(3):653--664, 2017.

\bibitem{devlin2012dynamic}
Sam~Michael Devlin and Daniel Kudenko.
\newblock Dynamic potential-based reward shaping.
\newblock In {\em Proceedings of the 11th International Conference on
  Autonomous Agents and Multiagent Systems}, pages 433--440. IFAAMAS, 2012.

\bibitem{espeholt2018impala}
Lasse Espeholt, Hubert Soyer, Remi Munos, Karen Simonyan, Volodymir Mnih, Tom
  Ward, Yotam Doron, Vlad Firoiu, Tim Harley, Iain Dunning, et~al.
\newblock Impala: Scalable distributed deep-rl with importance weighted
  actor-learner architectures.
\newblock {\em arXiv preprint arXiv:1802.01561}, 2018.

\bibitem{fortunato2017noisy}
Meire Fortunato, Mohammad~Gheshlaghi Azar, Bilal Piot, Jacob Menick, Ian
  Osband, Alex Graves, Vlad Mnih, Remi Munos, Demis Hassabis, Olivier Pietquin,
  et~al.
\newblock Noisy networks for exploration.
\newblock {\em arXiv preprint arXiv:1706.10295}, 2017.

\bibitem{Francois_2017}
Vincent François-Lavet.
\newblock {\em Contributions to deep reinforcement learning and its
  applications in smartgrids}.
\newblock PhD thesis, Universit\'e de Li\`ege, Li\`ege, Belgique, Sept. 2017.

\bibitem{fu2020autogan}
Yonggan Fu, Wuyang Chen, Haotao Wang, Haoran Li, Yingyan Lin, and Zhangyang
  Wang.
\newblock Autogan-distiller: Searching to compress generative adversarial
  networks.
\newblock {\em arXiv preprint arXiv:2006.08198}, 2020.

\bibitem{fujimoto2018addressing}
Scott Fujimoto, Herke Van~Hoof, and David Meger.
\newblock Addressing function approximation error in actor-critic methods.
\newblock {\em arXiv preprint arXiv:1802.09477}, 2018.

\bibitem{Gandhi_2017}
D. {Gandhi}, L. {Pinto}, and A. {Gupta}.
\newblock Learning to fly by crashing.
\newblock In {\em 2017 IEEE/RSJ International Conference on Intelligent Robots
  and Systems (IROS)}, pages 3948--3955, 2017.

\bibitem{gomez2005evolving}
Faustino Gomez and J{\"u}rgen Schmidhuber.
\newblock Evolving modular fast-weight networks for control.
\newblock In {\em International Conference on Artificial Neural Networks},
  pages 383--389. Springer, 2005.

\bibitem{haarnoja2018soft}
Tuomas Haarnoja, Aurick Zhou, Pieter Abbeel, and Sergey Levine.
\newblock Soft actor-critic: Off-policy maximum entropy deep reinforcement
  learning with a stochastic actor.
\newblock {\em arXiv preprint arXiv:1801.01290}, 2018.

\bibitem{hasselt2010double}
Hado Hasselt.
\newblock Double q-learning.
\newblock {\em Advances in neural information processing systems},
  23:2613--2621, 2010.

\bibitem{he2020milenas}
Chaoyang He, Haishan Ye, Li Shen, and Tong Zhang.
\newblock Milenas: Efficient neural architecture search via mixed-level
  reformulation.
\newblock In {\em Proceedings of the IEEE/CVF Conference on Computer Vision and
  Pattern Recognition}, pages 11993--12002, 2020.

\bibitem{he2016deep}
Kaiming He, Xiangyu Zhang, Shaoqing Ren, and Jian Sun.
\newblock Deep residual learning for image recognition.
\newblock In {\em Proceedings of the IEEE conference on computer vision and
  pattern recognition}, pages 770--778, 2016.

\bibitem{he2016identity}
Kaiming He, Xiangyu Zhang, Shaoqing Ren, and Jian Sun.
\newblock Identity mappings in deep residual networks.
\newblock In {\em European conference on computer vision}, pages 630--645.
  Springer, 2016.

\bibitem{henderson2017deep}
Peter Henderson, Riashat Islam, Philip Bachman, Joelle Pineau, Doina Precup,
  and David Meger.
\newblock Deep reinforcement learning that matters.
\newblock {\em arXiv preprint arXiv:1709.06560}, 2017.

\bibitem{hessel2017rainbow}
Matteo Hessel, Joseph Modayil, Hado Van~Hasselt, Tom Schaul, Georg Ostrovski,
  Will Dabney, Dan Horgan, Bilal Piot, Mohammad Azar, and David Silver.
\newblock Rainbow: Combining improvements in deep reinforcement learning.
\newblock {\em arXiv preprint arXiv:1710.02298}, 2017.

\bibitem{hester2017deep}
Todd Hester, Matej Vecerik, Olivier Pietquin, Marc Lanctot, Tom Schaul, Bilal
  Piot, Dan Horgan, John Quan, Andrew Sendonaris, Gabriel Dulac-Arnold, et~al.
\newblock Deep q-learning from demonstrations.
\newblock {\em arXiv preprint arXiv:1704.03732}, 2017.

\bibitem{hinton2015distilling}
Geoffrey Hinton, Oriol Vinyals, and Jeff Dean.
\newblock Distilling the knowledge in a neural network.
\newblock {\em arXiv preprint arXiv:1503.02531}, 2015.

\bibitem{howard2019searching}
Andrew Howard, Mark Sandler, Grace Chu, Liang-Chieh Chen, Bo Chen, Mingxing
  Tan, Weijun Wang, Yukun Zhu, Ruoming Pang, Vijay Vasudevan, et~al.
\newblock Searching for mobilenetv3.
\newblock In {\em Proceedings of the IEEE International Conference on Computer
  Vision}, pages 1314--1324, 2019.

\bibitem{hu2020dsnas}
Shoukang Hu, Sirui Xie, Hehui Zheng, Chunxiao Liu, Jianping Shi, Xunying Liu,
  and Dahua Lin.
\newblock Dsnas: Direct neural architecture search without parameter
  retraining.
\newblock In {\em Proceedings of the IEEE/CVF Conference on Computer Vision and
  Pattern Recognition}, pages 12084--12092, 2020.

\bibitem{hutter2019automated}
Frank Hutter, Lars Kotthoff, and Joaquin Vanschoren.
\newblock {\em Automated machine learning: methods, systems, challenges}.
\newblock Springer Nature, 2019.

\bibitem{jang2016categorical}
Eric Jang, Shixiang Gu, and Ben Poole.
\newblock Categorical reparameterization with gumbel-softmax.
\newblock {\em arXiv preprint arXiv:1611.01144}, 2016.

\bibitem{johnson2016perceptual}
Justin Johnson, Alexandre Alahi, and Li Fei-Fei.
\newblock Perceptual losses for real-time style transfer and super-resolution.
\newblock In {\em European conference on computer vision}, pages 694--711.
  Springer, 2016.

\bibitem{kang2018policy}
Bingyi Kang, Zequn Jie, and Jiashi Feng.
\newblock Policy optimization with demonstrations.
\newblock In {\em International Conference on Machine Learning}, pages
  2469--2478, 2018.

\bibitem{kim2013learning}
Beomjoon Kim, Amir-massoud Farahmand, Joelle Pineau, and Doina Precup.
\newblock Learning from limited demonstrations.
\newblock In {\em Advances in Neural Information Processing Systems}, pages
  2859--2867, 2013.

\bibitem{konda2000actor}
Vijay~R Konda and John~N Tsitsiklis.
\newblock Actor-critic algorithms.
\newblock In {\em Advances in neural information processing systems}, pages
  1008--1014, 2000.

\bibitem{lee2020journey}
Royson Lee, {\L}ukasz Dudziak, Mohamed Abdelfattah, Stylianos~I Venieris, Hyeji
  Kim, Hongkai Wen, and Nicholas~D Lane.
\newblock Journey towards tiny perceptual super-resolution.
\newblock {\em arXiv preprint arXiv:2007.04356}, 2020.

\bibitem{Levine_2016}
Sergey Levine, Chelsea Finn, Trevor Darrell, and Pieter Abbeel.
\newblock End-to-end training of deep visuomotor policies.
\newblock {\em J. Mach. Learn. Res.}, 17(1):1334–1373, Jan. 2016.

\bibitem{lillicrap2015continuous}
Timothy~P Lillicrap, Jonathan~J Hunt, Alexander Pritzel, Nicolas Heess, Tom
  Erez, Yuval Tassa, David Silver, and Daan Wierstra.
\newblock Continuous control with deep reinforcement learning.
\newblock {\em arXiv preprint arXiv:1509.02971}, 2015.

\bibitem{liu2019auto}
Chenxi Liu, Liang-Chieh Chen, Florian Schroff, Hartwig Adam, Wei Hua, Alan~L
  Yuille, and Li Fei-Fei.
\newblock Auto-deeplab: Hierarchical neural architecture search for semantic
  image segmentation.
\newblock In {\em Proceedings of the IEEE conference on computer vision and
  pattern recognition}, pages 82--92, 2019.

\bibitem{liu2018darts}
Hanxiao Liu, Karen Simonyan, and Yiming Yang.
\newblock Darts: Differentiable architecture search.
\newblock {\em arXiv preprint arXiv:1806.09055}, 2018.

\bibitem{liu2017stein}
Yang Liu, Prajit Ramachandran, Qiang Liu, and Jian Peng.
\newblock Stein variational policy gradient.
\newblock {\em arXiv preprint arXiv:1704.02399}, 2017.

\bibitem{mnih2016asynchronous}
Volodymyr Mnih, Adria~Puigdomenech Badia, Mehdi Mirza, Alex Graves, Timothy
  Lillicrap, Tim Harley, David Silver, and Koray Kavukcuoglu.
\newblock Asynchronous methods for deep reinforcement learning.
\newblock In {\em International conference on machine learning}, pages
  1928--1937, 2016.

\bibitem{mnih2015human}
Volodymyr Mnih, Koray Kavukcuoglu, David Silver, Andrei~A Rusu, Joel Veness,
  Marc~G Bellemare, Alex Graves, Martin Riedmiller, Andreas~K Fidjeland, Georg
  Ostrovski, et~al.
\newblock Human-level control through deep reinforcement learning.
\newblock {\em nature}, 518(7540):529--533, 2015.

\bibitem{nair2018overcoming}
Ashvin Nair, Bob McGrew, Marcin Andrychowicz, Wojciech Zaremba, and Pieter
  Abbeel.
\newblock Overcoming exploration in reinforcement learning with demonstrations.
\newblock In {\em 2018 IEEE International Conference on Robotics and Automation
  (ICRA)}, pages 6292--6299. IEEE, 2018.

\bibitem{ng1999policy}
Andrew~Y Ng, Daishi Harada, and Stuart Russell.
\newblock Policy invariance under reward transformations: Theory and
  application to reward shaping.
\newblock In {\em ICML}, volume~99, pages 278--287, 1999.

\bibitem{parisotto2015actor}
Emilio Parisotto, Jimmy~Lei Ba, and Ruslan Salakhutdinov.
\newblock Actor-mimic: Deep multitask and transfer reinforcement learning.
\newblock {\em arXiv preprint arXiv:1511.06342}, 2015.

\bibitem{pham2018efficient}
Hieu Pham, Melody~Y Guan, Barret Zoph, Quoc~V Le, and Jeff Dean.
\newblock Efficient neural architecture search via parameter sharing.
\newblock {\em arXiv preprint arXiv:1802.03268}, 2018.

\bibitem{Pinto_2017}
Lerrel Pinto, Marcin Andrychowicz, Peter Welinder, Wojciech Zaremba, and Pieter
  Abbeel.
\newblock Asymmetric actor critic for image-based robot learning.
\newblock 06 2018.

\bibitem{polino2018model}
Antonio Polino, Razvan Pascanu, and Dan Alistarh.
\newblock Model compression via distillation and quantization.
\newblock {\em arXiv preprint arXiv:1802.05668}, 2018.

\bibitem{real2019regularized}
Esteban Real, Alok Aggarwal, Yanping Huang, and Quoc~V Le.
\newblock Regularized evolution for image classifier architecture search.
\newblock In {\em Proceedings of the aaai conference on artificial
  intelligence}, volume~33, pages 4780--4789, 2019.

\bibitem{recht2019tour}
Benjamin Recht.
\newblock A tour of reinforcement learning: The view from continuous control.
\newblock {\em Annual Review of Control, Robotics, and Autonomous Systems},
  2:253--279, 2019.

\bibitem{rummery1994line}
Gavin~A Rummery and Mahesan Niranjan.
\newblock {\em On-line Q-learning using connectionist systems}, volume~37.
\newblock University of Cambridge, Department of Engineering Cambridge, UK,
  1994.

\bibitem{rusu2015policy}
Andrei~A Rusu, Sergio~Gomez Colmenarejo, Caglar Gulcehre, Guillaume Desjardins,
  James Kirkpatrick, Razvan Pascanu, Volodymyr Mnih, Koray Kavukcuoglu, and
  Raia Hadsell.
\newblock Policy distillation.
\newblock {\em arXiv preprint arXiv:1511.06295}, 2015.

\bibitem{schaal1997learning}
Stefan Schaal.
\newblock Learning from demonstration.
\newblock In {\em Advances in neural information processing systems}, pages
  1040--1046, 1997.

\bibitem{schaul2015prioritized}
Tom Schaul, John Quan, Ioannis Antonoglou, and David Silver.
\newblock Prioritized experience replay.
\newblock {\em arXiv preprint arXiv:1511.05952}, 2015.

\bibitem{schmitt2018kickstarting}
Simon Schmitt, Jonathan~J Hudson, Augustin Zidek, Simon Osindero, Carl Doersch,
  Wojciech~M Czarnecki, Joel~Z Leibo, Heinrich Kuttler, Andrew Zisserman, Karen
  Simonyan, et~al.
\newblock Kickstarting deep reinforcement learning.
\newblock {\em arXiv preprint arXiv:1803.03835}, 2018.

\bibitem{schulman2015trust}
John Schulman, Sergey Levine, Pieter Abbeel, Michael Jordan, and Philipp
  Moritz.
\newblock Trust region policy optimization.
\newblock In {\em International conference on machine learning}, pages
  1889--1897, 2015.

\bibitem{schulman2017proximal}
John Schulman, Filip Wolski, Prafulla Dhariwal, Alec Radford, and Oleg Klimov.
\newblock Proximal policy optimization algorithms.
\newblock {\em arXiv preprint arXiv:1707.06347}, 2017.

\bibitem{silver2014deterministic}
David Silver, Guy Lever, Nicolas Heess, Thomas Degris, Daan Wierstra, and
  Martin Riedmiller.
\newblock Deterministic policy gradient algorithms.
\newblock 2014.

\bibitem{stamoulis2019single}
Dimitrios Stamoulis, Ruizhou Ding, Di Wang, Dimitrios Lymberopoulos, Bodhi
  Priyantha, Jie Liu, and Diana Marculescu.
\newblock Single-path nas: Designing hardware-efficient convnets in less than 4
  hours.
\newblock In {\em Joint European Conference on Machine Learning and Knowledge
  Discovery in Databases}, pages 481--497. Springer, 2019.

\bibitem{sutton1988learning}
Richard~S Sutton.
\newblock Learning to predict by the methods of temporal differences.
\newblock {\em Machine learning}, 3(1):9--44, 1988.

\bibitem{sutton2000policy}
Richard~S Sutton, David~A McAllester, Satinder~P Singh, and Yishay Mansour.
\newblock Policy gradient methods for reinforcement learning with function
  approximation.
\newblock In {\em Advances in neural information processing systems}, pages
  1057--1063, 2000.

\bibitem{tan2019mnasnet}
Mingxing Tan, Bo Chen, Ruoming Pang, Vijay Vasudevan, Mark Sandler, Andrew
  Howard, and Quoc~V Le.
\newblock Mnasnet: Platform-aware neural architecture search for mobile.
\newblock In {\em Proceedings of the IEEE Conference on Computer Vision and
  Pattern Recognition}, pages 2820--2828, 2019.

\bibitem{tan2019efficientnet}
Mingxing Tan and Quoc~V Le.
\newblock Efficientnet: Rethinking model scaling for convolutional neural
  networks.
\newblock {\em arXiv preprint arXiv:1905.11946}, 2019.

\bibitem{teh2017distral}
Yee Teh, Victor Bapst, Wojciech~M Czarnecki, John Quan, James Kirkpatrick, Raia
  Hadsell, Nicolas Heess, and Razvan Pascanu.
\newblock Distral: Robust multitask reinforcement learning.
\newblock In {\em Advances in Neural Information Processing Systems}, pages
  4496--4506, 2017.

\bibitem{thrun1993issues}
Sebastian Thrun and Anton Schwartz.
\newblock Issues in using function approximation for reinforcement learning.
\newblock In {\em Proceedings of the 1993 Connectionist Models Summer School
  Hillsdale, NJ. Lawrence Erlbaum}, 1993.

\bibitem{van2015deep}
Hado Van~Hasselt, Arthur Guez, and David Silver.
\newblock Deep reinforcement learning with double q-learning.
\newblock {\em arXiv preprint arXiv:1509.06461}, 2015.

\bibitem{van2011insights}
Hado~Philip van Hasselt.
\newblock {\em Insights in reinforcement learning}.
\newblock Hado van Hasselt, 2011.

\bibitem{vecerik2017leveraging}
Mel Vecerik, Todd Hester, Jonathan Scholz, Fumin Wang, Olivier Pietquin, Bilal
  Piot, Nicolas Heess, Thomas Roth{\"o}rl, Thomas Lampe, and Martin Riedmiller.
\newblock Leveraging demonstrations for deep reinforcement learning on robotics
  problems with sparse rewards.
\newblock {\em arXiv preprint arXiv:1707.08817}, 2017.

\bibitem{wan2020fbnetv2}
Alvin Wan, Xiaoliang Dai, Peizhao Zhang, Zijian He, Yuandong Tian, Saining Xie,
  Bichen Wu, Matthew Yu, Tao Xu, Kan Chen, et~al.
\newblock Fbnetv2: Differentiable neural architecture search for spatial and
  channel dimensions.
\newblock In {\em Proceedings of the IEEE/CVF Conference on Computer Vision and
  Pattern Recognition}, pages 12965--12974, 2020.

\bibitem{wang2016sample}
Ziyu Wang, Victor Bapst, Nicolas Heess, Volodymyr Mnih, Remi Munos, Koray
  Kavukcuoglu, and Nando de Freitas.
\newblock Sample efficient actor-critic with experience replay.
\newblock {\em arXiv preprint arXiv:1611.01224}, 2016.

\bibitem{wang2016dueling}
Ziyu Wang, Tom Schaul, Matteo Hessel, Hado Hasselt, Marc Lanctot, and Nando
  Freitas.
\newblock Dueling network architectures for deep reinforcement learning.
\newblock In {\em International conference on machine learning}, pages
  1995--2003, 2016.

\bibitem{watkins1992q}
Christopher~JCH Watkins and Peter Dayan.
\newblock Q-learning.
\newblock {\em Machine learning}, 8(3-4):279--292, 1992.

\bibitem{wiewiora2003principled}
Eric Wiewiora, Garrison~W Cottrell, and Charles Elkan.
\newblock Principled methods for advising reinforcement learning agents.
\newblock In {\em Proceedings of the 20th International Conference on Machine
  Learning (ICML-03)}, pages 792--799, 2003.

\bibitem{wu2019fbnet}
Bichen Wu, Xiaoliang Dai, Peizhao Zhang, Yanghan Wang, Fei Sun, Yiming Wu,
  Yuandong Tian, Peter Vajda, Yangqing Jia, and Kurt Keutzer.
\newblock Fbnet: Hardware-aware efficient convnet design via differentiable
  neural architecture search.
\newblock In {\em Proceedings of the IEEE Conference on Computer Vision and
  Pattern Recognition}, pages 10734--10742, 2019.

\bibitem{wu2017scalable}
Yuhuai Wu, Elman Mansimov, Roger~B Grosse, Shun Liao, and Jimmy Ba.
\newblock Scalable trust-region method for deep reinforcement learning using
  kronecker-factored approximation.
\newblock In {\em Advances in neural information processing systems}, pages
  5279--5288, 2017.

\bibitem{xie2018snas}
Sirui Xie, Hehui Zheng, Chunxiao Liu, and Liang Lin.
\newblock Snas: stochastic neural architecture search.
\newblock {\em arXiv preprint arXiv:1812.09926}, 2018.

\bibitem{You_2017}
Yurong You, Xinlei Pan, Ziyan Wang, and Cewu Lu.
\newblock Virtual to real reinforcement learning for autonomous driving.
\newblock {\em CoRR}, abs/1704.03952, 2017.

\bibitem{zhang2018pretraining}
Xiaoqin Zhang and Huimin Ma.
\newblock Pretraining deep actor-critic reinforcement learning algorithms with
  expert demonstrations.
\newblock {\em CoRR}, abs/1801.10459, 2018.

\bibitem{zhu2020transfer}
Zhuangdi Zhu, Kaixiang Lin, and Jiayu Zhou.
\newblock Transfer learning in deep reinforcement learning: A survey.
\newblock {\em arXiv preprint arXiv:2009.07888}, 2020.

\bibitem{zoph2016neural}
Barret Zoph and Quoc~V Le.
\newblock Neural architecture search with reinforcement learning.
\newblock {\em arXiv preprint arXiv:1611.01578}, 2016.

\bibitem{zoph2018learning}
Barret Zoph, Vijay Vasudevan, Jonathon Shlens, and Quoc~V Le.
\newblock Learning transferable architectures for scalable image recognition.
\newblock In {\em Proceedings of the IEEE conference on computer vision and
  pattern recognition}, pages 8697--8710, 2018.

\end{thebibliography}
}

\appendix

\section{Evaluation of the scalability with model sizes on more tasks}
\label{appendix:scalability}

We evaluate the scalability of DRL with different model sizes on more tasks to better support the analysis in Sec.~\ref{sec:scalability}. As shown in Tab.~\ref{tab_appendix:scalability}, we can draw observations consistent with Sec.~\ref{sec:scalability} that \underline{(1)} larger model sizes with a higher network capability generally benefit the achieved test scores especially on more complex games, and \underline{(2)} there exists a task-specific optimal model size where larger models cannot further improve or even degrade the test score, which motivates the task-specific agent design.

\begin{table}[h]
\caption{The best test scores achieved by different models on more Atari games.}
\label{tab_appendix:scalability}
\begin{threeparttable}
\centering
\resizebox{1\linewidth}{!}{
{
\begin{tabular}{c||ccccc}
\toprule
\textbf{Atari Games} & \textbf{Vanilla} & \textbf{ResNet-14} & \textbf{ResNet-20} & \textbf{ResNet-38} & \textbf{ResNet-74} \\ \midrule
Breakout & 523.7 & 776.5 & 811 & \textbf{818.5} & 2.2 \\ 
Alien & 1724 & 9007 & \textbf{9323} & 8829 & 4456 \\ 
Asterix & 4850 & 708500 & \textbf{856800} & 756120 & 539060 \\ 
Atlantis & 3064320 & 3127390 & 3156130 & \textbf{3181090} & 3046490 \\ 
TimePilot & 4780 & 9070 & \textbf{9680} & 9500 & 9040 \\ 
SpaceInvaders & 1171 & 9848 & \textbf{46870} & 17962 & 15111 \\ 
WizardOfWor & 1320 & 2690 & \textbf{3580} & 3160 & 1850 \\ 
Qbert & 15085 & 13587 & 12385 & \textbf{15577.5} & 14097 \\ 
Pong & -19.9 & \textbf{21.0} & 20.9 & 20.9 & 20.8 \\
Tennis & -23.7 & 13.8 & 11.5 & \textbf{19.6} & 19.3 \\ 
Amidar & 721.8 & 638.6 & \textbf{845.2} & 138.3 & 493.9 \\ 
Asteroids & 2095 & 5690 & \textbf{5744} & 1947 & 4792 \\
Assault & 10164 & 14470 & \textbf{17314} & 12406.5 & 9849 \\
BankHeist & 1152 & \textbf{1288} & 1284 & 981 & 1163 \\
BattleZone & 7600 & 5800 & 13100 & \textbf{13300} & 4100 \\ 
BeamRider & 5530 & 23984 & 25961 & 29498 & \textbf{30048} \\ 
Bowling & 28.1 & 53 & \textbf{59.2} & 33.2 & 50.8 \\ 
Boxing & 4.2 & \textbf{100} & \textbf{100} & 99.3 & 87.1 \\
Centipede & 5025 & 6690 & 6410 & 6384.6 & \textbf{6899} \\ 
ChopperCommand & 1320 & 11170 & \textbf{14910} & 4370 & 8240 \\ 
CrazyClimber & 118300 & 128710 & 129550 & 130620 & \textbf{132720} \\ 
DemonAttack & 318349 & 481818 & 484382 & \textbf{494569} & 448450 \\ 
Gopher & 11914 & \textbf{63926} & 51112 & 42720 & 15340 \\
Gravitar & 475 & 565 & 490 & \textbf{660} & 490 \\ 
Jamesbond & 515 & 570 & 575 & \textbf{640} & 580 \\ 
Kangaroo & 160 & \textbf{8950} & 120 & 120 & 140 \\ 
Krull & 9412 & \textbf{9466.2} & 7020 & 7360.2 & 9192.3 \\ 
KungFuMaster & \textbf{35830} & 35450 & 34390 & 34750 & 34240 \\ 
MsPacman & 2712 & 4541 & \textbf{5853} & 3676 & 2556 \\ 
NameThisGame & 5808 & 17784 & \textbf{18908} & 18135 & 14409 \\ 
PrivateEye & \textbf{100} & \textbf{100} & \textbf{100} & \textbf{100} & \textbf{100} \\ 
Riverraid & 9164 & \textbf{31757} & 29037 & 27146 & 23521 \\ \bottomrule
\end{tabular}
}
}
\end{threeparttable}
\vspace{-1em}
\end{table}

\section{Evaluation of the proposed distillation mechanism on more tasks}
\label{appendix:distillation}

To further validate the proposed distillation mechanism in Sec.~\ref{sec:distillation}, we benchmark the proposed distillation mechanism with the SOTA policy distillation method~\cite{rusu2015policy} for training the vanilla network and ResNet-14 on more tasks as shown in Tab.~\ref{tab_appendix:distillation_policy}. We can observe that our proposed distillation achieves higher test scores in 26 out of the total 32 cases and comparable test scores in the remaining cases, verifying the superiority of distilling both the actor and critic as analyzed in Sec.~\ref{sec:distillation}.

\section{ResNet-based network structure}
\label{appendix:network}
The overall network design follows existing AC-based DRL methods~\cite{espeholt2018impala, mnih2016asynchronous}, which use a heavy feature extractor and two light-headers (implemented using two fully-connected layers) to design the actor and critic, respectively. For the feature extractor part, we follow the ResNet structure for CIFAR-10 dataset in~\cite{he2016identity} with two modifications: (1) we modify the stride of the first convolution to be two in order to adapt to the 84$\times$84 input resolutions of Atari games, and (2) we modify the output dimension of the final fully-connected layer to be 256, which serves as the input to the actor and critic headers.

\section{A2D's search space}
\label{appendix:supernet}

The supernet structure of A2D follows the \#groups, \#channels, and stride settings as the ResNet-like manual network design in Appendix.~\ref{appendix:network}. In particular, the supernet contains three groups with 5, 4, 5 searchable blocks respectively and all the blocks within the same group share the same basic channel numbers. Inspired by FBNet~\cite{wu2019fbnet}, we have nine candidate operations for each searchable block: standard convolutions with a kernel size 3/5, inverted residual blocks with a kernel size 3/5, a channel expansion of 1/3/5, and skip connections, leading to a search space of $9^{14}$ choices.

\begin{table*}[thb]
\vspace{-1em}
\caption{Benchmark the proposed distillation mechanism with (1) training without distillation, and (2) training with policy distillation~\cite{rusu2015policy} without distilling the critic.}
\label{tab_appendix:distillation_policy}
\begin{threeparttable}
\centering
\resizebox{1\textwidth}{!}{
{
\begin{tabular}{c|ccc|ccc}
\toprule
\multirow{3}{*}{\textbf{Atari Games}}& \multicolumn{3}{c|}{ \textbf{Vanilla} } & \multicolumn{3}{c}{ \textbf{ResNet-14} } \\
 & No & Policy  & \textbf{The proposed} & No & Policy & \textbf{The proposed} \\
  & distillation & distillation~\cite{rusu2015policy}  & \textbf{distillation} & distillation & disitllation~\cite{rusu2015policy} & \textbf{distillation} \\\midrule
Atlantis & 3064320 & 3159240 & \textbf{3160470} & 3127390 & 2845630 & \textbf{3148450} \\ 
Alien & 1724 & 3096 & \textbf{3419} & 9007 & 14682 & \textbf{15723} \\ 
TimePilot & 4780 & 9800 & \textbf{10200} & 9070 & 10490 & \textbf{10510} \\
SpaceInvaders & 1171 & 26821 & \textbf{39274} & 9848 & 76246 & \textbf{111189} \\ 
WizardOfWor & 1320 & \textbf{6310} & 5960 & 2690 & \textbf{6300} & 5450 \\
Asterix & 4850 & 59020 & \textbf{64510} & 708500 & 749870 & \textbf{849400} \\ 
Qbert & 15085 & \textbf{15687.5} & 15625 & 13587 & \textbf{17980} & 15470 \\ 
Amidar & 721.8 & \textbf{1452.5} & 1340.6 & 638.6 & 1439 & \textbf{1450.2} \\ 
Asteroids & 2095 & 4131 & \textbf{4647} & 5690 & 15371 & \textbf{15947} \\ 
BattleZone & 7600 & 12700 & \textbf{14500} & 5800 & 16300 & \textbf{18200} \\ 
BeamRider & 5530 & 14417 & \textbf{17806} & 23984 & 38311 & \textbf{42365} \\ 
Boxing & 4.2 & 2.8 & \textbf{100} & \textbf{100} & \textbf{100} & \textbf{100} \\ 
Centipede & 5025 & 5800 & \textbf{6575.5} & 6690 & 7744.3 & \textbf{8056.9} \\ 
ChopperCommand & 1320 & 15900 & \textbf{19120} & 11170 & 26320 & \textbf{31190} \\ 
CrazyClimber & 118300 & 138610 & \textbf{145700} & 128710 & 135290 & \textbf{138470} \\ 
DemonAttack & 318349 & 463823 & \textbf{483490} & 481818 & 517801 & \textbf{521051} \\ \bottomrule
\end{tabular}
}
}
\end{threeparttable}
\vspace{-1em}
\end{table*}


\end{document}